# Prostate Segmentation from Ultrasound Images using Residual Fully Convolutional Network

Md Sazzad Hossain
Faculty of Information Technology
Monash University.
Melbourne, Australia.
sazzad.hossain@monash.edu

Andrew P. Paplinski
Faculty of Information Technology
Monash University.
Melbourne, Australia.
sazzad.hossain@monash.edu

John M. Betts
Faculty of Information Technology
Monash University.
Melbourne, Australia.
sazzad.hossain@monash.edu

***Abstract*—Medical imaging based prostate cancer diagnosis procedure uses intra-operative transrectal ultrasound (TRUS) imaging to visualize the prostate shape and location to collect tissue samples. Correct tissue sampling from prostate requires accurate prostate segmentation in TRUS images. To achieve this, this study uses a novel residual connection based fully convolutional network. The advantage of this segmentation technique is that it requires no pre-processing of TRUS images to perform the segmentation. Thus, it offers a faster and straightforward prostate segmentation from TRUS images. Results show that the proposed technique can achieve around 86% Dice Similarity accuracy using only few TRUS datasets.**

***Keywords—Prostate, TRUS images, semantic segmentation, deep neural network.***

## I. Introduction

Prostate cancer diagnosis procedure requires real-time transrectal ultrasound (TRUS) imaging of prostate to collect tissue samples to investigate [1]. Besides, proper treatment of prostate cancer requires brachytherapy where a small radioactive source is directly implanted into prostate tissue guided by TRUS imaging [2]. In both of these operations, the prostate region needs to be segmented to make treatment plan and estimate the dose distribution. At present, prostate segmentation in TRUS is performed semiautomatically or manually, i.e. they require partial or total human involvement [3-5]. The reason is that TRUS images contain heavy speckle noises where prostate boundary can hardly be recognized. So typical pixel/voxel intensity based segmentation methods cannot achieve the desired accuracy without some manual guidance. However, such partial/total manual intervention for prostate segmentation is very time consuming and cumbersome. Therefore, it is still now an open challenge to perform fully automatic prostate segmentation from TRUS.

Various methodologies have been proposed so far in the last few decades. Many of those focused on developing effective filtering techniques to reduce the speckle noises and improve the contrast in order to facilitate the edge detection performance. For example, Pathak et al. [6] applied “sticks” algorithm to reduce noise and improve contrast. Then they applied anisotropic diffusion filter to smoothen the noise-reduced TRUS images. Liu et al. [7] used histogram normalisation and radial bas-relief method to make prostate edge more distinguishable for segmentation. Kwoh et al. [8] used Fourier transformation based harmonics method to reduce spurious edges and thus smoothen the boundary. Abolmaesumi et al. [9] used sticks filters to enhance the prostate boundary and then applied probabilistic data association filter (PDAF) for the segmentation task. Sahba et al. [10] used morphological gray level transformation to make the prostate boundary more distinguishable than the background. Then they applied Kalman filtering and fuzzy inference system to segment the prostate. However, the primary limitation of filtering and edge based techniques is that they can hardly handle shadow artifacts that very commonly appears in ultrasound images. So, apart from the edge detection method, model based techniques have been applied in many studies, where prior shape knowledge about prostate was utilised. Gong et al. [11] used deformable superellipses to model the prostate shape and then applied a Bayesian segmentation algorithm. Shen et al. [4] used statistical shape models of the prostate to guide their active shape model based deformable segmentation technique. Ghanei et al. [12] applied 3D deformable surface model of the prostate instead of 2D shape models. The limitation of model based techniques is that the models iteratively try to fit the prostate boundary which is time-consuming. Besides, they require strong edge information in order to fit to the prostate shape, which reverts back to the disadvantages associated with edge detection techniques.

In recent years, few studies used convolutional neural networks (CNN) for the prostate segmentation. Anas et al. [13] used a U-net CNN which takes the TRUS image at one end and provides the binary segmentation at the other end. Such segmentation is also known as semantic segmentation [14]. Zeng et al. [5] used a similar U-net architecture with different number of convolution layers. They also used statistical shape models from Magnetic Resonance Images (MRI) to improve the segmentation accuracy. The key advantage of CNN based segmentation is that they require almost no pre-processing and perform segmentation on statistical basis rather than edge detection or deformable shape basis. Such statistical/probabilistic segmentation is particularly beneficial for TRUS images since these images are heavily loaded with noises and shadow artifacts. This is why this study has also been inspired to use CNN for the prostate segmentation from TRUS images.

In this study, a new end-to-end CNN (also known as fully convolutional network) has been proposed which uses remote and neighbouring residual connections between its layers. This architecture has previously been used in a different study by the same authors [15]. In that study, this network showed a significant accuracy for prostate segmentation from MRI images. Thereby, authors hypothesize that the same network would be effective for TRUS images as well. The key advantage of residual connections is that they make the convolution layers avoid the vanishing gradient problem and prevents accuracy degradation at deeper layers as explained in [16]. Although residual connections have been previously used in conventional CNNs for classification tasks [17-19], to authors' best knowledge, the proposed residual connection design in fully convolutional networks is still unreported in any other studies and hence it implies the main contribution of this study.

The following sections describes the proposed methodologies in details and discusses results qualitatively as well as quantitatively.

## II. MATERIALS

The TRUS dataset used in this study were collected from Alfred Hospital, Melbourne, which consists of volumetric images of only 4 patients. The image dataset were obtained through proper ethical clearance from both Alfred hospital and Monash University.

The volumetric images were converted into stack of 2D images. In this way, total 55 2D images were obtained from all the volumetric images. Each of the 2D images were in grayscale version with size 224 x 224 to match the desired input image of VGG19 CNN. These images were in portable network graphics (PNG) format to avoid compression loss as well as obtain the original quality of the image.

After the extraction of 2D images, each of them were manually segmented by expert to generate the binary segmentation version of those images. Since this study follows supervised learning method, a pair of original image and its binary segmentation version were used for both training and testing.

## III. METHODOLOGY

### A. Convolutional Neural Network

As plenty of books and tutorials have already discussed convolutional neural network (CNN) in details, only the core idea will be put into context in this study.

CNN can be expressed as the following feedforward operation:

$$U = convT(X, W) + b \; ; \; Z = \max(U, 0) \quad (1)$$

where $X$ is the input image of size $M$ x $N$ x $D_x$, W is a filter of size $P$ x $Q$ x $D_x$, $convT(.)$ indicates multidimensional convolution operation that produces a tensor $U$ of size $M$ x $N$ x $D_w$ and $b$ is the bias value added to all element of the convolution output. The output of the convolution layer $U$ enters the activation layer $Z$ whose output becomes the input for the next layer. The output of $Z$ is also known as a 'feature map', because filters are trained to extract various features from the input image. Convolution layers are equipped with pooling layers, which basically reduce the dimensionality of the feature maps for faster training. Among different pooling methods, maxpooling is the most popular one, which breaks down the feature maps into small subregions of size $c$ x $c$ and then picks up the highest value from each subregion. That is, maxpooling gathers the local maxima values in order to select the most important part of the feature maps and reduce the learnable parameters at the same time. A series of convolution layers is followed by one or a few fully connected layers and a classification layer, which resembles the architecture of a conventional multilayer perceptron (MLP). During learning phase, all the weights in the network are adjusted by backpropagation formula so that a particular class of image produces a desired statistical output.

### B. The VGG19 model

Simonyan and Zisserman proposed a very deep neural network model in 2014 named 'VGG16'. This model consists of 3 stacked convolution layers and two fully connected layers. Each of the stacked convolution layers were composed of several convolution layers followed by one maxpooling layer. Each of the fully connected layers contains 4096 neurons. The network ends with a softmax layer following the last fully connected layer, which generates the final classification output of 10 pre-defined classes.

The authors then proposed another similar model in the same study named 'VGG19'. This model is basically a small modification to VGG16 model that has just three extra convolution layers than VGG16. It was found that VGG19 outperformed VGG16 in classification proving the effectiveness of the extra layers. Although several other deep neural network models have been proposed over the years e.g. AlexNet, ResNet and InceptionNet, VGG19 model remained popular for image classification tasks due to its simpler architecture compared to other models and faster training time. Most importantly, VGG models have been significantly used in semantic segmentation tasks than any other models.

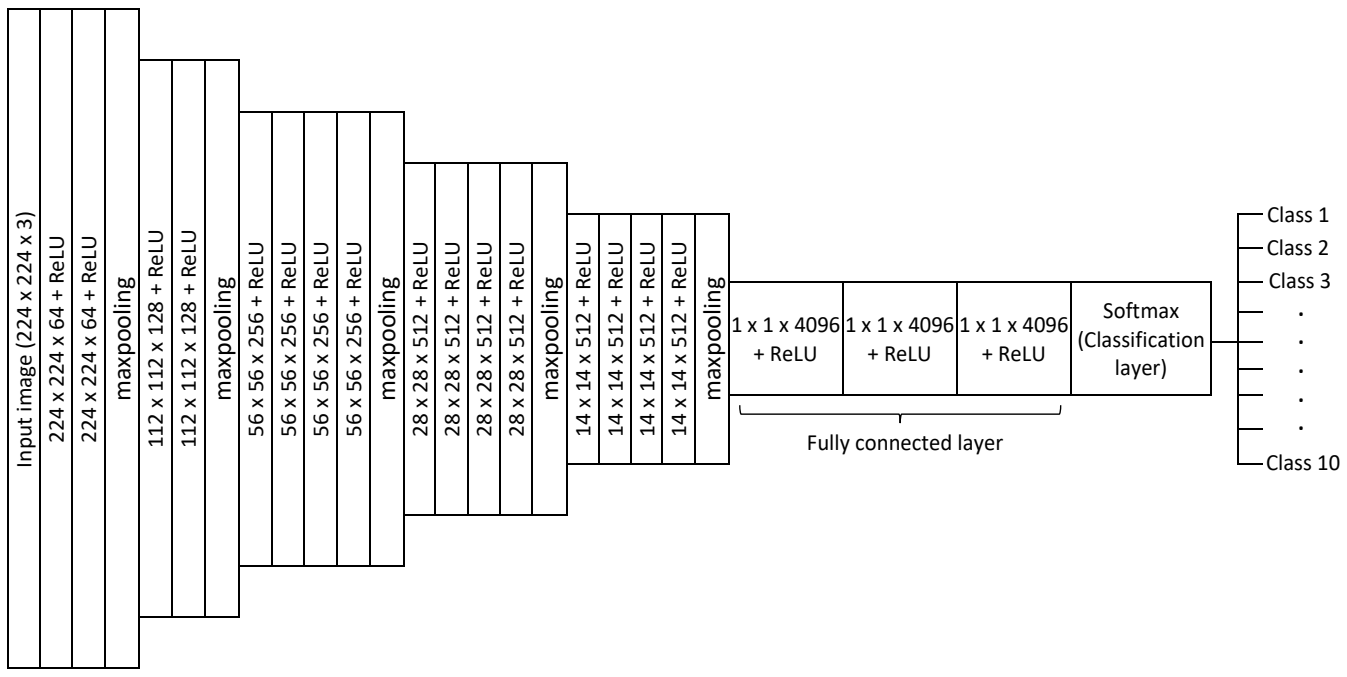

Fig. 1. The VGG19 model.

Fig. 1 illustrates the architecture of VGG19 CNN. It produces the classification output at the softmax layer, e.g. cat,

dog, car etc. It has 19 weight layers which serves the naming of 'VGG19'. These weight layers are the only part of the network that contains trainable parameters. VGG19 contains 16 convolution layers and two fully connected layers. Most of the convolution layers are stacked serially having only five maxpooling layers in between. Therefore, the convolution layers can be divided into 5 subregions. Convolution layers in each subregion generates same number of feature maps $D_w$. The number of feature maps in these subregions, $D_w$ = 64, 128, 256, 512 and 512 respectively. The output of final convolution layers is flattened and feed into the fully connected layers. The output of fully connected layers then enters the softmax layer which provides the ultimate probabilistic output on the input image such as 97% cat and 0.3% dog.

### *C. Fully convolutional network and semantic segmentation*

Conventional CNN was designed to classify one whole image. In contrast to this, the idea of semantic segmentation is that CNN will classify every pixel of the image and thus will provide an output image of classwise segmentations.

Semantic segmentation requires architectural modification of conventional CNN. This modifications mainly includes replacement of the fully connected layer with a 'deconvolution' part as shown in Fig. 2. This 'deconvolution' part is basically the mirror version of the convolution part. That is, it takes the feature maps from the last convolution layer and then by deconvolution and upsampling it outputs a class-segmented version of the original image. The deconvolution part is followed by a softmax layer which ultimately generates the semantically segmented image by performing pixelwise probabilistic classification. Fig. 2 shows an example of semantic segmentation by FCN where the input image contains a human figure and the output image contains its semantic segmentation version where human figure's region was separated by gray colour.

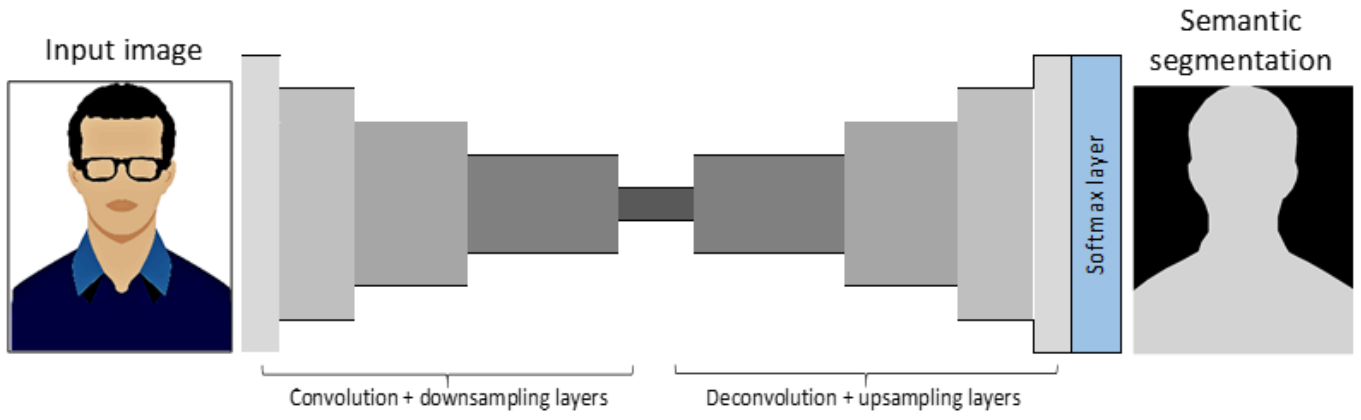

Fig. 2. Semantic segmentation by fully convolutional network (FCN).

### *D. VGG19RSeg: the proposed FCN model*

The proposed FCN model in this study uses the architecture of VGG19. To make the FCN model, the fully connected layer in VGG19 was replaced with a mirror version of VGG19's convolution part, which fulfils the requirement of 'deconvolution' part for the FCN. The final layer is the softmax layer to do the pixelwise classification. Apart from convolution, activation (ReLU) and maxpooling layer, batch normalization layer has also been added before each ReLU layer. Batch normalization helps to achieve quicker convergence by reducing the covariate shift [20].

In addition to the general architecture, this network uses two types of residual/skip connection – remote and neighbouring residual connection. Hence, the proposed name is VGG19RSeg. It has been firstly reported in one of our recent papers published in 2018 [15]. The remote residual connections has been also recently proposed in two other architectures, namely, U-net [21] and V-net [22]. Remote residual connection (RRC) basically connects the maxpooling layers of the convolution part and their corresponding max-un-pooling, i.e. upsampling layers of the deconvolution part. The advantage of such connections is that they facilitate the upsampling operation at deconvolution part by providing/copying the information of the feature maps from the corresponding maxpooling layer in the convolution part. However, the usage of both neighbouring residual connections (NRC) and RRC in VGG19 based FCN is yet to be reported in any other similar studies except our own previous study [15]. This neighbouring residual connection basically connects the first and last convolution layer of a group of stacked convolution layers. Such connection was inspired by the ResNet CNN structure proposed by He et al. [16], which won the first place in the ImageNet competition, 2015 proving its effectiveness over other CNN models. Primarily, residual connection helps the optimization procedure by diminishing the vanishing gradient problem that occurs in a deeper layer. Mathematically, if x and F(x) be the input and output of a convolution layer respectively, residual connection converts the output as F(x)+x. So if F(x) be zero due to vanishing gradient, the identity mapping x would still remain. Therefore, residual connection prevents increase of training error in a deeper layer than its shallower adjoined layer.

Since the input images are 2D monochromatic TRUS images, the initial number of channels has been reduced to 1 in the VGG19 architecture as shown in Fig. 3. Each input image was of size 224 x 224 to match the original input size for VGG19. The output image is a binary segmentation of original image, where white colour was assigned for the prostate region and black for the background, i.e. everything else.

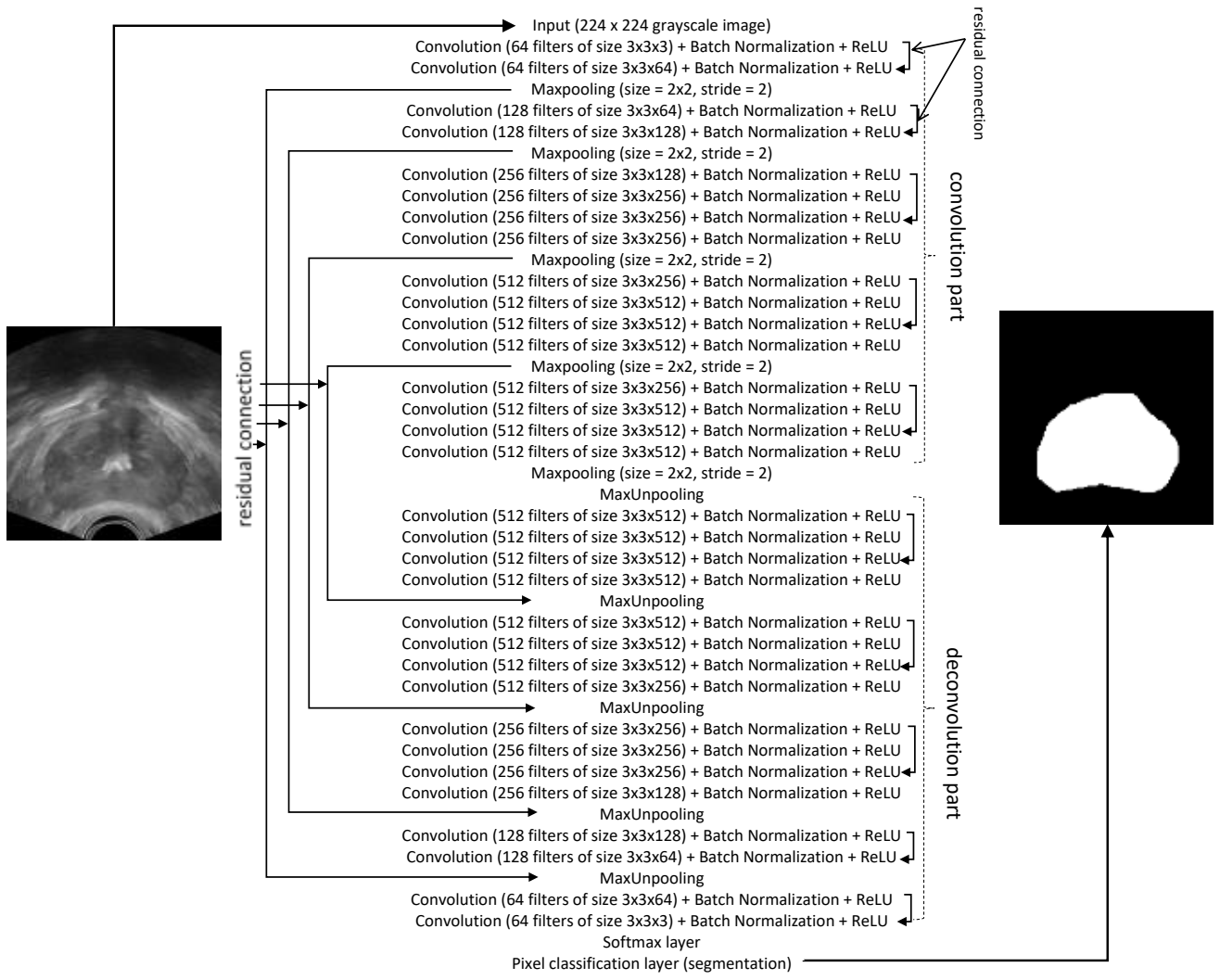

Fig. 3. The proposed FCN model – VGG19RSeg.

### E. Training scheme

As stated before, TRUS volumetric images were converted into 2D grayscale images of size 224 x 224. Due to lack of data availability, 90% of the 2D images and their respective manually segmented images were used for training and the rest 10% were used for testing purpose. To facilitate the training process, class-weighted cross entropy loss function was used similar to [23] expressed in Eq. (2). This function is particularly useful when there is lack of enough dataset and pixels of different classes are notably imbalanced, i.e. not equal in images.

$$L = -\frac{1}{n}\sum_{i=1}^{N} {w_i}^{cl}[\hat{P}_i log P_i + (1-\hat{P}_i)\log(1-P_i)] \quad (2)$$

$$\text{where, } {w_i}^{cl} = \frac{1}{pixels\ of\ class\ x_i}$$

In Eq. (2), $L$ represents the loss function with $P_i$ and $\hat{P}_i$ being the pixel of a given class marked by FCN and human expert respectively.

The training of FCN was carried out by a Monash University high performance cluster computing system named "Massive HPC". This computing system consists of 13 processors, 120GB memory and Nvidia Tesla K80 GPU. As for the training parameters, initial learning rate – 0.0005, momentum – 0.9, mini batch size – 4 and number of epochs was set at 30.

## IV. Results

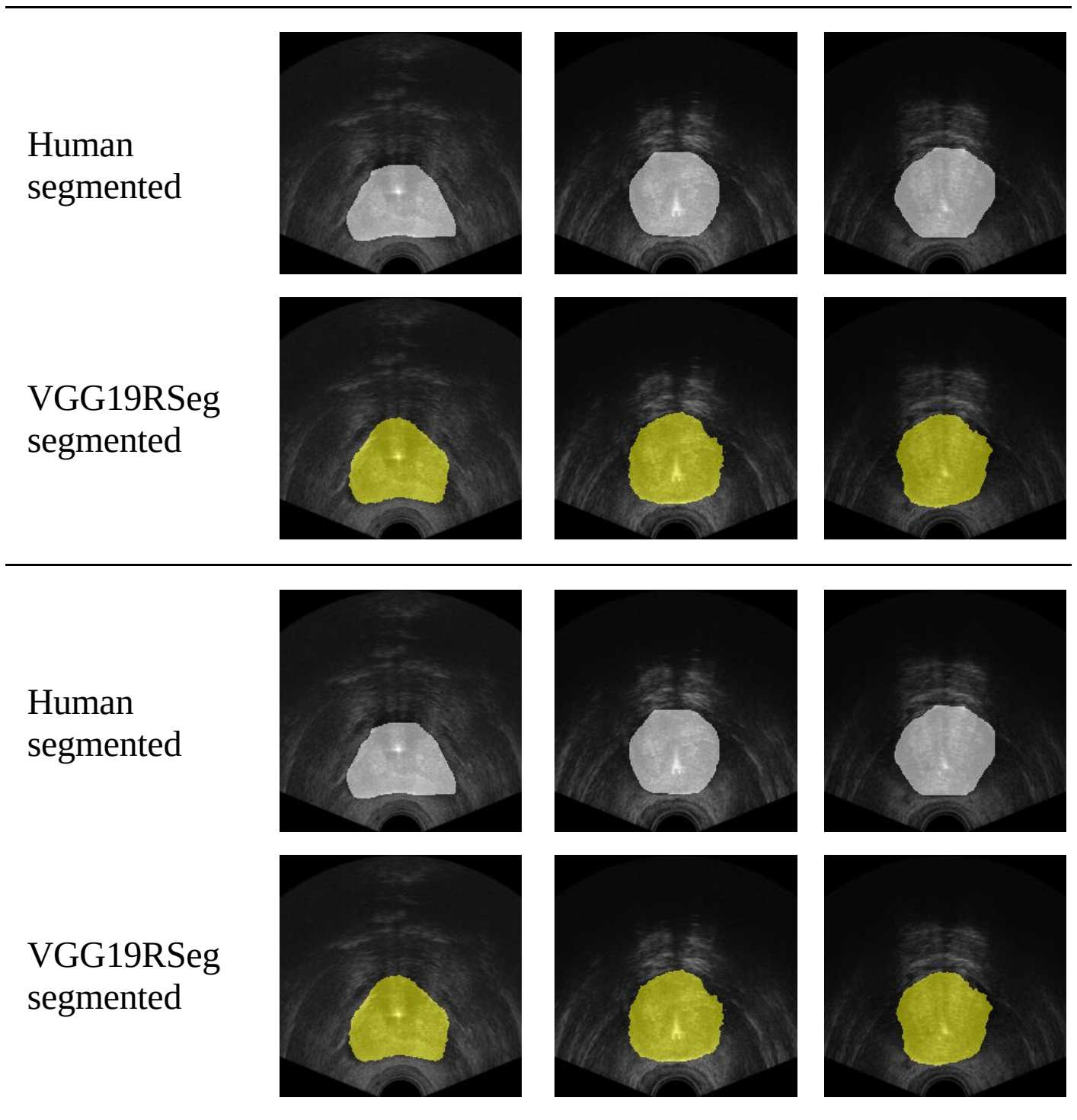

Fig. 4. Qualitative comparison between VGG19RSeg and manual segmentations.

The total training time was approximately 5 hours. The training accuracy reached as high as 97.38% at the end of the training. The performance of the proposed network was evaluated with test datasets which were 5 in total due to lack of dataset as mentioned before. Fig. 4 shows the qualitative comparison between human segmented and VGG19RSeg segmented image. It can be seen that VGG19RSeg provided very accurate segmentation that nearly matches the manual segmentations.

The accuracy was quantified using Dice Similarity Coefficient (DSC) as given in Eq. (3).

$$DSC = 2\frac{|X \cap Y|}{|X| + |Y|} \quad (3)$$

where X and Y indicates two separate regions/classes, where one of the regions is prostate region and the other one is the background. The modulus sign '| |' indicates the cardinal of the corresponding set.

Apart from the proposed model, another model – VGG-FCN has also been tested whose layer-by-layer architecture is same as VGG19RSeg, but does not have the NRC. The deployment of this model is to reveal the effect of NRC in an FCN. A third model – U-Net has been tested as well which have been previously used in a similar study on prostate segmentation in TRUS [5]. All three models were trained and tested 5 times with different randomly selected training and testing dataset each time to obtain a consistent result. Table I-III present the quantified accuracy in percentage of DSC of the VGG19-FCN, VGG19RSeg and U-Net respectively for all 5 observations.

TABLE I. ACCURACY OF VGG19RSEG WITHOUT NRC

| | DSC (%) | | | | |
|---|---|---|---|---|---|
| | Test run 1 | Test run 2 | Test run 3 | Test run 4 | Test run 5 |
| Average | 71.19 | 71.57 | 74.81 | 81.76 | 73.94 |
| Maximum | 94.57 | 95.85 | 89.24 | 89.56 | 92.31 |
| Minimum | 23.43 | 25.38 | 52.48 | 76.05 | 24.25 |

TABLE II. ACCURACY OF VGG19RSEG

| | DSC (%) | | | | |
|---|---|---|---|---|---|
| | Test run 1 | Test run 2 | Test run 3 | Test run 4 | Test run 5 |
| Average | 87.26 | 85.86 | 84.56 | 86.38 | 87.62 |
| Maximum | 94.41 | 95.98 | 99.72 | 94.20 | 94.16 |
| Minimum | 75.92 | 69.04 | 63.09 | 72.42 | 78.86 |

TABLE III. ACCURACY OF U-NET

| | DSC (%) | | | | |
|---|---|---|---|---|---|
| | Test run 1 | Test run 2 | Test run 3 | Test run 4 | Test run 5 |
| Average | 83.43 | 76.97 | 82.51 | 79.85 | 81.49 |
| Maximum | 90.72 | 93.55 | 93.20 | 92.69 | 94.53 |
| Minimum | 29.38 | 53.85 | 52.29 | 39.92 | 59.83 |

Table II shows that the average DSC of VGG19RSeg with NRC is 86.34%, whereas it's only 74.65% for the same model without NRC as shown in Table I. So, inclusion of NRC increased the accuracy by 12% approximately. Besides, Table III also shows that U-Net, which does not have NRC, yielded less accuracy than VGG19RSeg with average 80.85% DSC. Therefore, these results clearly indicates that NRC results in higher accuracy to an FCN.

Table IV shows a comparison between the present study and some other similar studies.

TABLE IV. COMPARISON OF VGG19RSEG WITH OTHER STUDIES

| | DSC (%) | Method | Data volume |
|---|---|---|---|
| Present study | 86.34 | VGG19RSeg | 4 |
| Zeng et al. [5] | 90.10 | U-Net | 605 |
| Anas et al. [24] | 92.91 | ResUnet | 3892 |
| Yan et al. [25] | 91 | ALSS | 19 |
| Yang et al. [26] | 92.39 | BCRNN | 17 |

As revealed in Table IV, although VGG19RSeg achieved slightly lower accuracy than other similar studies, only 4 patients' data were used here whereas other studies used 17-3892 patients' dataset. Therefore, compared to dataset volume, the present study obtained a significant accuracy which is likely to be further improved with larger dataset, because Table I-III has shown that VGG19RSeg provides the best accuracy using the same dataset volume.

## V. CONCLUSION

This study applies fully convolutional network (FCN) based semantic segmentation technique to perform prostate region segmentation from TRUS images. A new FCN model has been proposed for the task which contains neighbouring residual connections (NRC) between its stacked convolution layers. It also contains remote residual connections which have been previously used in different studies. This model basically adopts the architecture of a popular CNN model – VGG19 and modifies it to make the FCN model. This is why this model has been named as VGG19RSeg. Here, very few TRUS dataset were used to train the model due to data scarcity. However, results showed that the proposed model could achieve around 86% Dice Similarity Coefficient (DSC) accuracy on average even with such small dataset. Although some previous studies achieved slightly higher accuracy than this study has, those studies were found to use large number of TRUS datasets for their models. So compared to dataset volume, the proposed model in this study provided a promising accuracy. Another result showed that the proposed FCN model's architecture without the NRC degrades the accuracy by around 14%. Hence it proves that NRC in FCN contributes to higher accuracy in semantic segmentation. Future study will put effort on collecting more TRUS dataset as well as effective architectural modification to the FCN model in order to improve the present accuracy level.

## ACKNOWLEDGMENT

The TRUS dataset was collected from Alfred Hospital, Melbourne. Both Monash University and Alfred Hospital provided prompt ethical approval for the project. Besides, a very special thanks to Massive™ HPC who provided a very high configuration computational unit to train and test the FCN models.